# Academic Network Representation via Prediction-Sampling Incorporated Tensor Factorization

Chunyang Zhang, Xin Liao and Hao Wu

*Abstract*—Accurate representation to an academic network is of great significance to academic relationship mining like predicting scientific impact. A Latent Factorization of Tensors (LFT) model is one of the most effective models for learning the representation of a target network. However, an academic network is often High-Dimensional and Incomplete (HDI) because the relationships among numerous network entities are impossible to be fully explored, making it difficult for an LFT model to learn accurate representation of the academic network. To address this issue, this paper proposes a Prediction-sampling-based Latent Factorization of Tensors (PLFT) model with two ideas: 1) constructing a cascade LFT architecture to enhance model representation learning ability via learning academic network hierarchical features, and 2) introducing a nonlinear activation-incorporated predicting-sampling strategy to more accurately learn the network representation via generating new academic network data layer by layer. Experimental results from the three real-world academic network datasets show that the PLFT model outperforms existing models when predicting the unexplored relationships among network entities.

## I. Introduction

An academic network describes various relationships among scholars, institutions, researchers, and publications in the academic world. Mining academic relationship like predicting scientific impact is the most important task in analyzing academic networks, which heavily depends on how to accurately represent the target academic network in low-dimensional space [1-6]. Generally, an academic network is high-dimensional and incomplete (HDI), that is to say, it consists of numerous different type of nodes (i.e., scholars, papers), but only a few links between nodes are observed (i.e., two authors write the same paper) [7-11]. Therefore, given its HDI property, how to accurately learn the low-dimensional representation of an academic network becomes a thorny issue to academic relationships mining.

To date, researchers have proposed a variety of representation learning models [12-25]. Among them, a latent factorization of tensors (LFT)-based model has already demonstrated a powerful representation learning ability for a given network[26-35]. It models a target network as a tensor, thus, the structure information of network is preserved fully. For example, a fine-grained regularized LFT model [36], A generalized Nesterov's accelerated LFT model [37], and a neural tensor factorization model [38]. In particular, an LFT-based representation learning model is built based on the known data of a tensor only [39-40]. As a result, the representation learning ability of LFT model relies on the tensor data density. However, due to the HDI property of academic network, the data density of corresponding tensor is typically extremely low, i.e., such tensor is an HDI tensor. Hence, to learn the accurate representation of an academic network, can we build a high-performance LFT-based representation learning model by increasing the data density of a target HDI tensor?

To answer it, this paper innovatively proposes a Prediction-sampling-based Latent Factorization of Tensors (PLFT) model. Its main idea is twofold: a) an cascade structure LFT model is design based on the deep forest principle [41-42], thus, the hierarchical latent feature of academic network hidden in HDI tensor can be learned sequentially and eventually aggregated into the low-dimensional representation of the network, and b) a nonlinear activation-incorporated predicting sampling strategy is adopted to generate synthetic data layer by layer for increasing the data density of the target HDI tensor. Therefore, the main contributions of this study include:

1) A PLFT model. It is able to learn the accurate representation of an HDI academic network via fusing prediction-sampling strategy into cascade LFT architecture.

2) Detailed algorithm analysis and design for the proposed PLFT. It provides detailed guidance for researchers to implement PLFT.

3) Empirical studies on real-world academic network datasets demonstrate that the PLFT model achieves highly accurate representation to HDI academic network than state-of-the-art models do.

Section II gives related work. Section III gives preliminaries. Section IV introduces the methodology. Section V illustrates the detail of the experiments. Section VI concludes the whole paper.

C. Y. Zhang, X. Liao and H. Wu are with the College of Computer and Information Science, Southwest University, Chongqing 400715, China, (email: cyzh24@163.com, lxchat26@gmail.com, haowuf@gamil.com).

## II. RELATED WORK

Academic network representation learning has gained considerable attention in recent years due to its significance in understanding complex structures [43-45]. Several approaches have been proposed to capture the multi-relational nature of academic networks, ranging from shallow methods to deep learning-based methods. Traditional shallow methods [46-51], such as matrix factorization and random walk-based models, aiming to preserve network structure while generating low-dimensional embeddings. For example, DeepWalk [52] applies random walks to generate sequences of nodes and uses a skip-gram model to learn node embeddings, while node2vec [53] extends this approach by introducing flexible random walk strategies to capture both local and global structural information. Another notable model, LINE [54] explicitly preserves first-order and second-order proximities in large-scale networks, making it well-suited for academic networks. Additionally, metapath2vec [55] is specifically designed for academic networks, utilizing meta-path-based random walks to incorporate domain-specific semantic information. Despite their efficiency, these shallow models struggle to capture high-order dependencies and complex interactions within academic networks. To address these limitations, deep learning-based methods have been introduced [56], leveraging more complex structures such as graph neural networks (GNNs) to enhance representation learning in academic networks. Heterogeneous graph attention networks [57-58] have been applied to academic networks for the purpose of understanding of these complex network structures. GATNE [59], which uses a neural network to perform representation learning on academic networks by modeling both the node attributes and the multi-relational structure of the network. HeRec [60] incorporates cascaded neural network mechanisms to learn representation of academic networks. Additionally, Heterogeneous graph convolutional networks [61-63] offer a robust framework for modeling complex relationships in academic networks. However, deep models tend to require large-scale data and suffer from scalability issues when dealing with large academic datasets.

A complementary approach to academic network representation is tensor decomposition, which models multi-relational data by factorizing high-order tensors [64-71]. Unlike graph-based methods that primarily focus on pairwise interactions, tensor decomposition can effectively capture higher-order relationships, such as the co-occurrence of authors, papers, and venues. One of the earliest models, RESCAL [72-74], employs bilinear factorization to model multi-relational data and has been applied to academic knowledge graphs for citation and author disambiguation. Tucker decomposition [75-76] further extends this approach by introducing a core tensor that captures latent interactions across different dimensions, making it suitable for analyzing evolving academic collaboration patterns. Additionally, CP decomposition [77-80] can offer a highly interpretable framework for modeling research trends. Although these tensor decomposition methods [81] can be used as representations of learning academic networks, their performance tends to be heavily dependent on known data density, i.e., low density may result in poor performance.

To our knowledge, few successful attempts have been made to extract useful information from extremely sparse academic networks with tensor decomposition methods, nor to effectively use this information for relational prediction. The PLFT proposed in this paper utilizes the properties of academic networks to model complex context information, while exploiting the ability of LFT model to learn effective representations.

## III. PRELIMINARIES

*Definition* 1. An HDI tensor: Considering three node sets *I*, *J* and relation set *K*, a tensor $\mathbf{Y}^{|I|\times|J|\times|K|}$ has each entry $y_{ijk}$ of its elements that indicates a link weight between pairs of nodes through a specific relationship. Then, Given **Y**'s known and unknown element set $\Lambda$ and $\Gamma$, **Y** is an HDI tensor if $|\Lambda|<<|\Gamma|$.

*Definition* 2. Rank-one tensor: $\mathbf{A}_r^{|I|\times|J|\times|K|}$ is formulated by the outer product of three latent feature vectors $\boldsymbol{u}_r, \boldsymbol{s}_r, \boldsymbol{t}_r$, which can be mathematically expressed as $\mathbf{A}_r = \boldsymbol{u}_r \circ \boldsymbol{s}_r \circ \boldsymbol{t}_r$.

Note that $\boldsymbol{u}_r, \boldsymbol{s}_r$ and $\boldsymbol{t}_r$ have lengths $|I|$, $|J|$, and $|K|$ respectively. Expanding on this, the detailed expression for each element $a_{ijk}$ in $\mathbf{A}_r$ is given by

$$a_{ijk} = u_{ir} s_{jr} t_{kr} \tag{1}$$

where $u_{ir}$, $s_{jr}$ and $t_{kr}$ represent single elements in $\boldsymbol{u}_r, \boldsymbol{s}_r$ and $\boldsymbol{t}_r$, respectively. As shown in Fig .1, with *R* rank-one tensors, which means $\{\mathbf{A}_r \mid r \in \{1, 2, \cdots, R\}\}$, we obtain the rank-*R* approximation of Y, denoted $\hat{\mathbf{Y}}$ as follows:

$$\hat{\mathbf{Y}} = \sum_{r=1}^{R} \mathbf{A}_r \tag{2}$$

where each element of $\hat{\mathbf{Y}}$ is formulated as:

$$\hat{y}_{ijk} = \sum_{r=1}^{R} u_{ir} s_{jr} t_{kr} \tag{3}$$

To achieve the latent feature (LF) matrices, this paper employs Euclidean distance [82] as a measure to quantify the difference between **Y** and $\hat{\mathbf{Y}}$. It is essential to incorporate a regularization term into the objective function to ensure the model's generalizability and prevent overfitting. The combined objective function is formulated as follows:

$$\varepsilon = \sum_{y_{ijk} \in \Lambda} \left( \left(\mathbf{Y} - \hat{\mathbf{Y}}\right)^2 + \lambda \left( \|\mathbf{U}\|_F^2 + \|\mathbf{S}\|_F^2 + \|\mathbf{T}\|_F^2 \right) \right) \quad (4)$$

where $\|.\|_F$ represents the Frobenius norm of a matrix. Then we focus on the loss incurred by each individual entry:

$$\varepsilon = \sum_{y_{ijk} \in \Lambda} \left( \left( y_{ijk} - \sum_{r=1}^R u_{ir} s_{jr} t_{kr} \right)^2 + \lambda \sum_{r=1}^R \left( u_{ir}^2 + s_{jr}^2 + t_{kr}^2 \right) \right) \quad (5)$$

$$s.t. \forall i \in I, j \in J, k \in K, r \in \{1,2,3,...,R\}: u_{ir} \geq 0, s_{ir} \geq 0, t_{ir} \geq 0$$

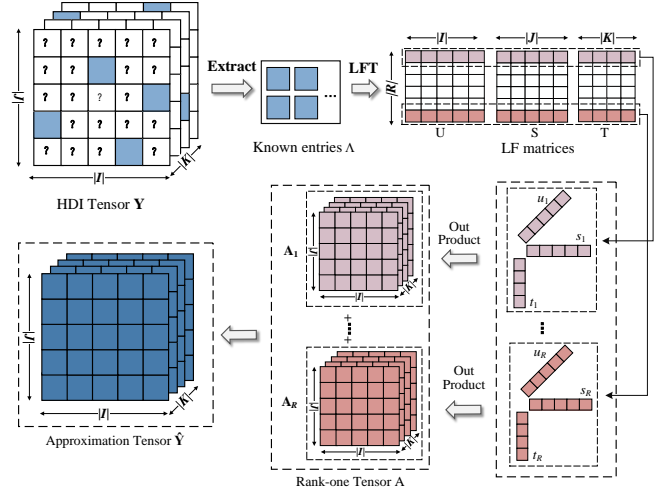

Fig. 1. Latent factorization of an HDI tensor **Y**.

## IV. METHODOLOGY

### A. Academic network HDI tensor

According to [83], academic network can be transformed into tensor within a multi-relational network framework. In this setup, each relationship links two specific-type entities, with their relationship type explicitly defined. A third-order tensor is used to represent multi-relational network, effectively encoding each element. The representation of academic network is modeled using a tensor **Y** with dimensions $|I| \times |J| \times |K|$, where each element encodes relationship between nodes $v_i$ and $v_j$ relative to the $k$-th relation. The tensor captures relational values along its third dimension, while two other dimensions represent the author nodes involved. The followings are illustrative examples of the significance of relation values. If $k$ is 1, it corresponds to two authors writing the same paper; if $k$ is 2, it corresponds to two authors writing separate papers, but both papers were published at the same conference; if $k$ is 3, it corresponds to two authors writing separate papers, but both papers use common terminologies. The general process is shown in the tensor construction section on the left in Fig. 2.

To quantify the correlation among various relations, the element $y_{ijk}$ are processed utilizing the Gaussian copula method [84]. Each element signifies the link weight between two author nodes, with a higher value indicating a closer relationship between them.

### B. Structure of PLFT

Following the deep forest methodology proposed by Zhou *et al.* [41], the PLFT model is devised, as depicted in the Fig. 2. The PLFT model sequentially cascades $N$ LFT models and $N$-1 activation functions, forming a cascaded prediction-sampling-based architecture. The PLFT model works as follows:

1) Inputting $\Lambda$ as the initial inputs of PLFT;

2) Initializing $n = 1$;

3) Layer 1: training LF matrices $U_1$, $S_1$ and $T_1$ based on $\Lambda$ and $\Omega$ to obtain $\hat{Y}_1$. Note that the Generated synthetic entry set $\Omega$ of $\overline{Y}$ is empty;

4) Layer 1: selecting sequentially the unknown data from $\hat{Y}_1$. Traversing iteratively each row of slice matrices gained from relation dimension, and deliberately selecting a missing entry located between two known entries. If a row in a slice matrix contains no known data, a blank entry is randomly selected to be included in generation set. (Note that this approach ensures that our model can efficiently select unknown data while also accounting for the density of data);

5) Layer 1: predicting $\hat{y}_{ijk}$ of the selected entry and input $\hat{y}_{ijk}$ into an activation function to get output $\overline{y}_{ijk}$;

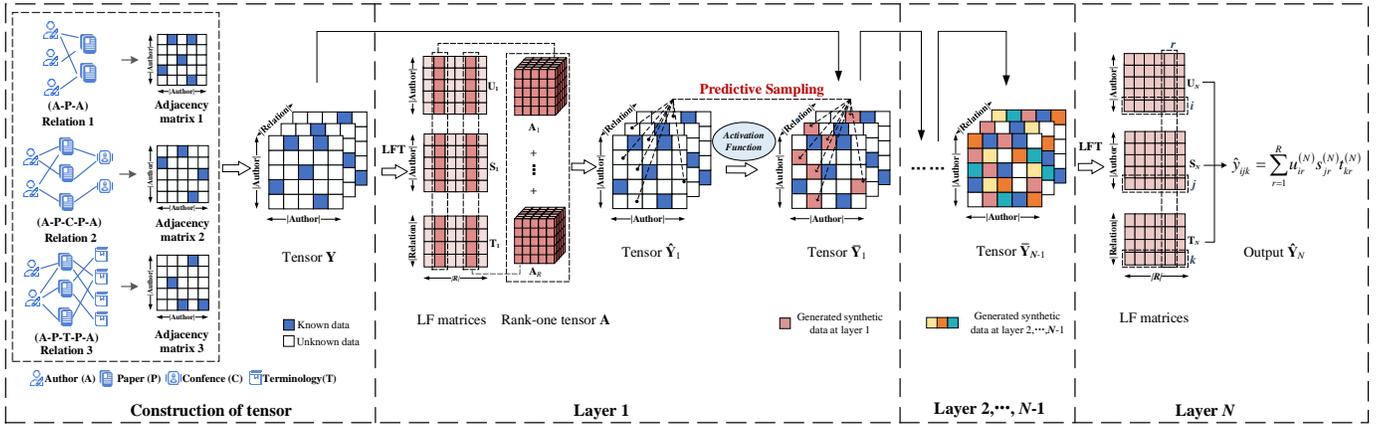

Fig. 2. The overall architecture diagram. It includes the construction process of tensor and cascaded prediction-sampling structure of the PLFT model

6) Layer 1: repeating steps 4-5 until the number of generated synthetic data sets Ω matches the number of known sets currently (the pre-set criterion). Subsequently, integrating newly generated entry sets from the current layer into the known data sets, and combining the outputs to form an HDI tensor $\overline{\mathbf{Y}}_1$;

7) Setting $n=n+1$;

8) Layer 2, ···, $N$-1: repeating steps 3–7 to ultimately obtain $\overline{\mathbf{Y}}_{N-1}$. Note that $n \in \{1, ···, N\}$;

9) Layer $N$: training $\mathbf{U}_N$, $\mathbf{S}_N$ and $\mathbf{T}_N$ to obtain $\hat{\mathbf{Y}}_N$, which can be used to formulate $\mathbf{Y}$'s final approximation;

Where $\overline{\mathbf{Y}}_{n-1}$ is the input HDI tensor of the $n$-th layer and $\hat{\mathbf{Y}}_n$ is the output HDI tensor of the $n$-th layer. The detailed process of generating synthetic data is shown in the predictive-sampling section in Fig. 2. Then a brief explanation is provided. Blank entries in each row are selected after slicing by relation, prioritizing blanks between known entities. $\mathbf{U}_n$, $\mathbf{S}_n$, and $\mathbf{T}_n$ are then used to predict values for these blanks, followed by applying an activation function to the predictions.

### C. Training the n-th Layer With Adam

To integrate the generated data from the $(n-1)$-th layer into the $n$-th layer during the training of the PLFT, this paper extends the formulation of the loss function from a single element to a more comprehensive form, considering the $n$-th layer as a general case where $n$ ranges from 1 to $N$. The following formula is annotated with relevant details.

$$\varepsilon(\mathbf{U}_n, \mathbf{S}_n, \mathbf{T}_n) = \underbrace{\sum_{\langle i,j,k \rangle \in \Lambda} \left( y_{ijk} - \sum_{r=1}^{R} u_{ir}^{(n)} s_{jr}^{(n)} t_{kr}^{(n)} \right)^2}_{\text{Training on known data}} + \alpha \underbrace{\sum_{\langle i,j,k \rangle \in \Omega} \left( \overline{y}_{ijk} - \sum_{r=1}^{R} u_{ir}^{(n)} s_{jr}^{(n)} t_{kr}^{(n)} \right)^2}_{\text{Training on generated synthetic data}} + \lambda \sum_{r=1}^{R} \left( \left(u_{ir}^{(n)}\right)^2 + \left(s_{jr}^{(n)}\right)^2 + \left(t_{kr}^{(n)}\right)^2 \right) \quad (6)$$

where $u_{ir}^{(n)}$, $s_{jr}^{(n)}$, $t_{kr}^{(n)}$ denote single elements of LF matrices at the $n$-th layer and $\overline{y}_{ijk}$ is a single element of Ω. $\alpha$ is a parameter that modulates the influence of the generated data in Ω.

$$\langle i,j,k \rangle \in \Lambda : \varepsilon_{ijk}^{(n)} = \left( y_{ijk} - \sum_{r=1}^{R} u_{ir}^{(n)} s_{jr}^{(n)} t_{kr}^{(n)} \right)^2 + \lambda \left( \sum_{r=1}^{R} \left(u_{ir}^{(n)}\right)^2 + \sum_{r=1}^{R} \left(s_{jr}^{(n)}\right)^2 + \sum_{r=1}^{R} \left(t_{kr}^{(n)}\right)^2 \right) \quad (7)$$

So the loss function of the first layer can be expressed as (6). Notably, when at the first layer (i.e., $n = 1$), there are only known entry sets, thus Ω remains empty.

Adam (Adaptive Moment Estimation) [85] is an advanced iterative optimization algorithm. It learns model parameters efficiently by adapting the learning rate for each parameter based on estimates of both the first moment and the second moment of the gradients.

Based on the update rules of Adam, the update rules for PLFT model is devised as follows. This is also divided into two cases on Λ and Ω. Note that the LF matrix $\mathbf{U}_n$ is used as an example, and the updates to the other LF matrices are similar.

$$\begin{cases} u_{ir}^{(n)} \leftarrow u_{ir}^{(n)} - \eta \dfrac{\hat{m}_u^{(x)}}{\sqrt{\hat{v}_u^{(x)}} + \tau} \\ \hat{m}_u^{(x)} = \dfrac{m_u^{(x)}}{1-\beta_1} \quad \hat{m}_u^{(x)} = \dfrac{v_u^{(x)}}{1-\beta_2} \end{cases} \quad (8)$$

where $x$ is the current number of iterations. $m_u^{(x)}$ and $v_u^{(x)}$ are the estimates of the first and second moments of the gradient, respectively. $\hat{m}_u^{(x)}$ and $\hat{v}_u^{(x)}$ are offset correction of $m_u^{(x)}$ and $v_u^{(x)}$. Then $m_u^{(x)}$ and $v_u^{(x)}$ are updated in two cases:

$$\begin{cases} \text{if } \langle i,j,k \rangle \in \Lambda, \ m_u^{(x)} = \beta_1 m_u^{(x-1)} + (1-\beta_1)\left( s_{jr}^{(n)} t_{kr}^{(n)} \left( y_{ijk} - \sum_{r=1}^{R} s_{jr}^{(n)} t_{kr}^{(n)} \right) - \lambda u_{ir}^{(n)} \right) \\ \text{if } \langle i,j,k \rangle \in \Omega, \ m_u^{(x)} = \beta_1 m_u^{(x-1)} + (1-\beta_1)\left( \alpha s_{jr}^{(n)} t_{kr}^{(n)} \left( y_{ijk} - \sum_{r=1}^{R} s_{jr}^{(n)} t_{kr}^{(n)} \right) - \lambda u_{ir}^{(n)} \right) \end{cases} \quad (9)$$

$$\begin{cases} \text{if } \langle i,j,k \rangle \in \Lambda, \ v_u^{(x)} = \beta_2 v_u^{(x-1)} + (1-\beta_2)\left( s_{jr}^{(n)} t_{kr}^{(n)} \left( y_{ijk} - \sum_{r=1}^{R} s_{jr}^{(n)} t_{kr}^{(n)} \right) - \lambda u_{ir}^{(n)} \right)^2 \\ \text{if } \langle i,j,k \rangle \in \Omega, \ v_u^{(x)} = \beta_2 v_u^{(x-1)} + (1-\beta_2)\left( \alpha s_{jr}^{(n)} t_{kr}^{(n)} \left( y_{ijk} - \sum_{r=1}^{R} s_{jr}^{(n)} t_{kr}^{(n)} \right) - \lambda u_{ir}^{(n)} \right)^2 \end{cases} \quad (10)$$

In the above formula, $\beta_1$ and $\beta_2$ are the exponential decay rates that govern the estimation of these two moments, and $\tau$ is a small constant that ensures numerical stability by avoiding division by zero. After $U_n$ is trained on the data, $T_n$ and $S_n$ are trained respectively by fixing the other matrices at the same epoch. Similarly, $m_s^{(x)}$, $m_t^{(x)}$, $v_s^{(x)}$, $v_t^{(x)}$ are the estimates of the first and second moments of the gradient, respectively. $\hat{m}_s^{(x)}$, $\hat{m}_t^{(x)}$, $\hat{v}_s^{(x)}$, $\hat{v}_t^{(x)}$ are offset correction of $m_s^{(x)}$, $m_t^{(x)}$, $v_s^{(x)}$ and $v_t^{(x)}$.

$$\begin{cases} s_{jr}^{(n)} \leftarrow s_{jr}^{(n)} - \eta \frac{\hat{m}_s^{(x)}}{\sqrt{\hat{v}_s^{(x)} + \tau}} \\ \hat{m}_s^{(x)} = \frac{m_s^{(x)}}{1-\beta_1} \quad \hat{m}_s^{(x)} = \frac{v_s^{(x)}}{1-\beta_2} \end{cases} \quad (11)$$

$$\begin{cases} t_{kr}^{(n)} \leftarrow t_{kr}^{(n)} - \eta \frac{\hat{m}_t^{(x)}}{\sqrt{\hat{v}_t^{(x)} + \tau}} \\ \hat{m}_t^{(x)} = \frac{m_t^{(x)}}{1-\beta_1} \quad \hat{m}_t^{(x)} = \frac{v_t^{(x)}}{1-\beta_2} \end{cases} \quad (12)$$

After training the LF matrices during each iteration, the final $U_n$, $S_n$ and $T_n$ are employed to predict missing value of the selected unknown data. Subsequently, the predicted values are evaluated within a specified threshold range to make a judgment. The mapping process adopts nonlinear activation function according to [39], which converts the predicted value $\hat{y}_{ijk}$ to $\bar{y}_{ijk}$.

$$\bar{y}_{ijk} = \begin{cases} y_{\min} + 1/\left(1+e^{-\hat{y}_{ijk}}\right), & \text{if } \hat{y}_{ijk} < y_{\min} \\ y_{\max} / \left(1+e^{-\hat{y}_{ijk}}\right), & \text{if } \hat{y}_{ijk} > y_{\max} \\ \hat{y}_{ijk}, & \text{otherwise} \end{cases} \quad (13)$$

Where the maximum and minimum values of **Y** are set to $y_{max}$ and $y_{min}$ respectively. The nonlinear activation function is employed to derive a new value $\bar{y}_{ijk}$ for predicted values that exceed $y_{max}$ or fall below $y_{min}$.

## V. EXPERIMENTAL RESULTS AND DISCUSSION

### A. General Settings

1) Datasets: The experiments are conducted on three datasets (cora[1], aminer[2], dblp[3]), and the details of these are summarized in Table II. Specifically, we utilize a subset of dblp, containing 4057 authors, 14328 papers, 7723 terms, and 20 conferences. Similarly, we use a subset of aminer, containing 4057 authors, 14328 papers, and 4 conferences. Furthermore, we leverage a subset of cora, containing 8052 authors, 20201 papers and 12313 terms. Note that dblp contains three types of relations (APA, APCPA and APTPA) in Fig. 2, while cora contains two relations(APA and APCPA) and aminer also contains two relations (APA and APTPA). In the experiments, each dataset is split into training set, validation set and test set by 80%, 10%, 10% randomly.

---

[1] http://www.cs.umd.edu/~sen/lbc-proj/LBC.html
[2] https://www.aminer.cn/data
[3] https://dblp.uni-trier.de

TABLE I. DATASETS DETAILS

| Dataset | cora | aminer | dblp |
|---|---|---|---|
| Author Count | 17411 | 8052 | 4057 |
| Relation Count | 2 | 2 | 3 |
| Element Count | 48954 | 33470 | 16131 |
| Density | 0.0080% | 0.0258% | 0.0326% |

Note that each academic network has initially been constructed with over ten million known entities. Due to the particularity of the cascade prediction-sampling structure, the network undergoes sparsification.

2) Evaluation Metrics: The representation learning ability of an LFT model for an HDI tensor can be assessed by measuring its accuracy in predicting missing values [86-87]. For such purposes, two metrics are adopted: Root Mean Square Error (RMSE) and Mean Absolute Error (MAE). They are calculated as follows:

$$\text{RMSE} = \sqrt{\sum_{y_{ijk} \in \psi} \left( y_{ijk} - \hat{y}_{ijk} \right)^2 / |\psi|} \qquad \text{MAE} = \sum_{y_{ijk} \in \psi} \left| y_{ijk} - \hat{y}_{ijk} \right| / |\psi|$$

where $\psi$ denotes the validation set. A low MAE and RMSE indicate high prediction accuracy, as they signify that the predicted values are close to the actual values with minimal deviations. All experiments are conducted on a PC with a 2.1GHz Intel Core i7-13700 CPU with 32GB of RAM.

B. Comparison Results

This paper undertakes a comparative analysis of the PLFT model against state-of-the-art models that excel in accurately predicting missing link weights between author nodes. The experimental scope encompasses an array of models, focusing on tensor-based approaches and non-tensor structural models.

(a) M1: A PLFT model proposed in this paper.
(b) M2: A CTF model proposed in [88]. It enhances the robustness to outliers by combining tensor decomposition and Cauchy loss.
(c) M3: A TCA model proposed in [89]. It adopts CP decomposition to construct multi-dimensional relations and effectively fill the missing values in the tensor.
(d) M4: A HCCF model proposed in [90]. It proposes a self-supervised recommendation framework to enhance the discriminant ability of GNN-based CF paradigms.
(e) M5: A SHT model proposed in [91]. It enhances the data of the interaction graph through cross-perspective self-supervised learning.
(f) M6: A MGDN model proposed in [92]. It treats graph neural networks as untrainable Markov processes, using distance weighting to construct features of vertices.

Model settings are standardized as follows: a) setting $R = 20$ for tensor-based models including PLFT, CTF and TCA; b) following original specifications for layer count and hidden unit dimensions for other models based on CF and GNN (HCCF, SHT and MGDN); c) setting $\alpha = 1.5$, $N = 10$ for PLFT; e) setting $\beta_1$ to 0.9, $\beta_2$ to 0.999, $\tau$ to 1e-8; f) optimizing regularization coefficients and learning rates for all models to achieve optimal predictive accuracy. Each model is trained for 1000 iterations or until the error between iterations is below $10^{-5}$. Models are evaluated five times, with average results reported for reliability and fairness.

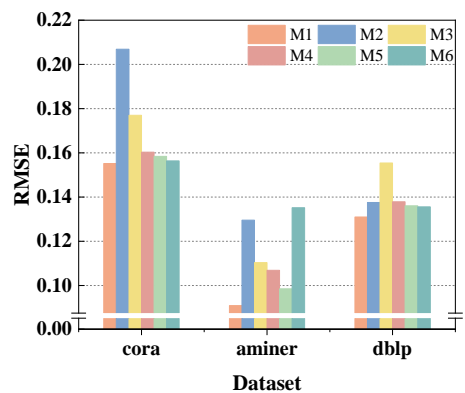

Fig. 3. The comparison results on prediction accuracy of the six models

*Comparison of Prediction Accuracy*: The comparative outcomes are summarized in Fig .3, revealing that in all three datasets, the RMSE and MAE values for PLFT model consistently outperform those of the state-of-the-art models. Specifically, on aminer, M1's

RMSE stands at 0.0909, which shows a reduction of 29.86%, 17.66%, 14.49%, 7.72%, and 32.76% compared to M2, M3, M4, M5, and M6 respectively. Additionally, M1's MAE of 0.0565 is 28.12% lower than M2's 0.0786, 12.40% lower than M3's 0.0645, 24.57% lower than M4's 0.0749, 17.28% lower than M5's 0.0683, and 54.80% lower than M6's 0.1250, demonstrating the model's robust ability to minimize errors in prediction. These patterns of improvement are also observable across the other datasets, with M1 demonstrating marginal advantages in some cases. The results confirm that PLFT model achieves significantly higher prediction accuracy compared to the benchmark models, underscoring its distinct advantage in accurately modeling and predicting incomplete HDI tensor data. The findings indicate that PLFT is not only effective but also reliable in diverse real-world scenarios, positioning it as a promising solution for prediction.

Furthermore, we conducted a Wilcoxon signed-rank test [93] to assess whether PLFT significantly outperforms every single model in predictive accuracy. The comparison results in Table III include three indices: $w+$, $w-$, $p$-value. A higher $w+$ value indicates greater prediction accuracy, while $p$-value reflects the level of statistical significance. The results clearly demonstrate that the prediction accuracy of PLFT is significantly superior to all the comparison models.

TABLE II. WILCOXON SIGNED-RANK TEST RESULTS ON RMSEs & MAEs

| Comparison | $w+$ | $w-$ | $p$-value* |
|---|---|---|---|
| M1 vs. M2 | 21 | 0 | **0.03125** |
| M1 vs. M3 | 21 | 0 | **0.03125** |
| M1 vs. M4 | 21 | 0 | **0.03125** |
| M1 vs. M5 | 21 | 0 | **0.03125** |
| M1 vs. M6 | 21 | 0 | **0.03125** |

* Hypotheses accepted at a 0.05 significance level are highlighted

## C. The effect of Cascaded Predictive Sampling

During these experiments, the impact of varying the number of layers on the prediction accuracy of PLFT are systematically evaluated. To ensure a fair comparison, consistent hyper-parameter settings are maintained across all tests, with $\lambda$ set to 0.01, $\eta$ to 0.001, and $R$ fixed at 20. As depicted in Fig. 4, the outcomes of PLFT reveal insightful trends across different layer configurations. From this analysis, following key observations are derived:

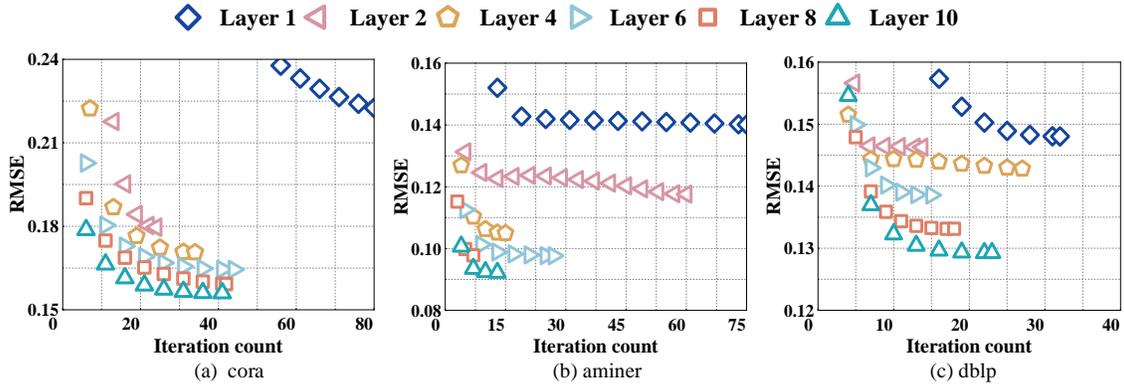

Fig. 4. The training process of PLFT regarding RMSE at different layers on all the datasets

1) The convergence of single layer within the prediction-sampling architecture of PLFT model remains unaffected, with RMSE/MAE consistently decreasing across training rounds until a stable convergence is achieved for each layer. In subsequent layers, synthetic data markedly speeds up convergence, requiring fewer iterations compared to initial layer. For example, on cora, the first layer converges after 258 rounds, while the fifth and tenth layers converge after merely 86 and 82 rounds, respectively. This trend is consistently observed across diverse datasets showing that PLFT's cascaded structure significantly reduces the required iterations for convergence.

2) The PLFT model, which is built upon a cascaded prediction-sampling structure, notably enhances the predictive accuracy of the LFT model. As the number of layers increases, a clear trend emerges where the RMSE and MAE metrics for PLFT decrease, signifying improved performance. Notably, when PLFT is configured with a single layer, it essentially reverts to the original LFT model. However, by increasing the number of layers, a substantial improvement in prediction accuracy is observed. For example, with just one layer, the RMSE/MAE stands at 0.2147/0.1584 on cora. However, upon introducing the second layer, there is a sharp decline in both RMSE to 0.1635 and MAE to 0.1237, with subsequent layers yielding only marginal improvements. These stress the benefit of the cascaded structure in enhancing predictive accuracy.

3) The prediction accuracy of PLFT model does not consistently scale with an increase in its depth. For example, on aminer, the RMSE initially decreases by 0.0920 up to the second layer, but subsequent layers show a slight increase. This case can be attributed to the introduction of potentially unreliable prediction sampling values into the subsequent layers once the optimal number of layers is surpassed. Nevertheless, it is noteworthy that even when the model is extended to the tenth layer, the resulting RMSE /MAE metric remain significantly lower than those observed in the single layer, highlighting overall effectiveness of the cascaded structure in sampling enhancing predictive performance .

## VI. CONCLUSION

This paper proposes a prediction-sampling-based latent factorization of tensors (PLFT) model, which effectively tackles the challenges posed by high-dimensional and incomplete academic networks. By leveraging a cascade LFT architecture and incorporating a nonlinear activation-based predicting-sampling approach, the PLFT model provides a more efficient solution for academic network representation through the extraction of hierarchical features and the generation of more precise representations. The PLFT model is evaluated across three real-world datasets, demonstrating its strong competitiveness in predicting hidden connections between network entities. Therefore, the PLFT model has the ability to solve the issue of uncovering unknown relationships in academic networks.

We plan to address these issues in future: enhancing the computational efficiency of PLFT via a parallel framework and making hyperparameter adaptive through intelligent algorithms. These improvements will further optimize the model's performance, enabling it to handle larger and more complex academic networks.